\PassOptionsToPackage{table}{xcolor}
\documentclass{article}

\usepackage{xcolor}

\usepackage[tracking=true]{microtype}
\SetTracking{encoding=*}{-10}
\usepackage{graphicx}
\usepackage{subcaption}
\usepackage{booktabs} 

\usepackage{hyperref}
\usepackage{xurl}

\usepackage[preprint]{icml2026}

\usepackage{amsmath}
\usepackage{amssymb}
\usepackage{mathtools}
\usepackage{amsthm}

\usepackage{tikz}
\usetikzlibrary{arrows, positioning, shapes.geometric, shadows, backgrounds, fit, arrows.meta, calc}
\usepackage{array}
\usepackage{tabularx}
\usepackage{algorithm}
\usepackage{algorithmic}

\usepackage[capitalize,noabbrev]{cleveref}

\usepackage{tcolorbox}
\tcbuselibrary{skins} 
\usepackage{multicol} 
\usepackage{enumitem} 
\theoremstyle{plain}
\newtheorem{theorem}{Theorem}[section]

\theoremstyle{definition}
\newtheorem{definition}[theorem]{Definition}

\theoremstyle{remark}

\usepackage[textsize=tiny]{todonotes}

\definecolor{FrontierBlue}{RGB}{25, 118, 210}
\definecolor{FrontierPurple}{RGB}{126, 87, 194}
\definecolor{FrontierSlate}{RGB}{71, 85, 105}
\definecolor{FrontierBG}{RGB}{248, 250, 252}
\definecolor{TableHeaderBG}{RGB}{235, 245, 255}
\definecolor{TableRowAlt}{RGB}{250, 251, 252}

\definecolor{LabGreenBG}{RGB}{230, 247, 235}
\definecolor{LabGreenBorder}{RGB}{56, 142, 60}
\definecolor{LabBlueBG}{RGB}{235, 245, 255}
\definecolor{LabBlueBorder}{RGB}{25, 118, 210}
\definecolor{LabPurple}{RGB}{126, 87, 194}
\definecolor{LabOrange}{RGB}{255, 152, 0}
\definecolor{LabSlate}{RGB}{33, 33, 33}

\icmltitlerunning{Position: Foundation Models for Tabular Data within Systemic Contexts Need Grounding}

\begin{document}

\twocolumn[
  \icmltitle{Position: Foundation Models for Tabular Data within Systemic Contexts Need Grounding}

  \icmlsetsymbol{equal}{*}

  \begin{icmlauthorlist}
    \icmlauthor{Tassilo Klein}{sap}
    \icmlauthor{Johannes Hoffart}{sap}
  \end{icmlauthorlist}

  \icmlaffiliation{sap}{SAP SE}
  \icmlcorrespondingauthor{Tassilo Klein}{tassilo.klein@sap.com}

  \icmlkeywords{Machine Learning, Tabular Data, Foundation Models, Semantically Linked Tables}

  \vskip 0.08in
]

\printAffiliationsAndNotice

\setlength{\parskip}{2pt}
\setlength{\parsep}{1pt}

\begin{abstract}
This position paper argues that foundation models for tabular data face inherent limitations when isolated from operational context---the procedural logic, declarative rules, and domain knowledge that define how data is created and governed. Current approaches focus on single-table generalization or schema-level relationships, fundamentally missing the operational knowledge that gives data meaning.We introduce Semantically Linked Tables (SLT) and Foundation Models for SLT (FMSLT) as a new model class that grounds tabular data in its operational context. We propose dual-phase training: pre-training on open-source code-data pairs and synthetic systems to learn business logic mechanics, followed by zero-shot inference on proprietary data. We introduce the ``Operational Turing Test'' benchmark and argue that operational grounding is essential for autonomous agents in complex data environments.
\end{abstract}

\section{Introduction}
\label{sec:intro}
\looseness=-1 Foundation models have demonstrated remarkable capacity to generalize across diverse tasks and datasets, extending beyond training data~\cite{bommasani2022opportunitiesrisksfoundationmodels,Huang2022LanguageMA}. However, applying these models to tabular data analysis is hindered by oversimplifying assumptions. Current work on foundation models for tabular data sometimes conflates distinct problems: some focus primarily on ``isolated tables''~\cite{pmlr-v235-van-breugel24a}, a legitimate perspective in certain scenarios but one that fails to reflect the realities of intricate, interconnected data ecosystems.
\begin{figure}[t]
    \centering
    \begin{tikzpicture}[
        font=\sffamily,
        node distance=0.3cm,
        scale=0.6, transform shape,
        execute at begin picture={
            \definecolor{LabGreenBG}{RGB}{230, 247, 235}      
            \definecolor{LabGreenBorder}{RGB}{56, 142, 60}    
            \definecolor{LabBlueBG}{RGB}{235, 245, 255}       
            \definecolor{LabBlueBorder}{RGB}{25, 118, 210}    
            \definecolor{LabPurple}{RGB}{126, 87, 194}        
            \definecolor{LabOrange}{RGB}{255, 152, 0}         
            \definecolor{LabSlate}{RGB}{33, 33, 33}           
        },
        inner_box/.style={
            draw=LabGreenBorder!35, 
            fill=white, 
            fill opacity=1.0,
            line width=0.5pt, 
            rounded corners=6pt, 
            inner sep=7pt, 
            align=left, 
            text width=5.5cm,
            drop shadow={shadow xshift=0.3mm, shadow yshift=-0.3mm, opacity=0.12}
        },
        green_container/.style={
            fill=LabGreenBG!60, 
            draw=LabGreenBorder!70, 
            line width=0.8pt, 
            rounded corners=12pt
        },
        blue_container/.style={
            fill=LabBlueBG!50, 
            fill opacity=0.55,
            draw=LabBlueBorder!70, 
            dashed, 
            dash pattern=on 4pt off 2pt,
            line width=0.9pt, 
            rounded corners=12pt
        }
    ]


    \node[inner_box, fill=white] (relational) {
        {\fontsize{8}{9.5}\selectfont\bfseries\color{LabSlate} Relational data}\\[1.2pt]
        {\fontsize{6.5}{8}\selectfont\color{LabSlate!85} Relational data, mostly database tables linked by foreign keys}
    };

    \node[inner_box, above=0.5cm of relational, fill=white] (operational) {
        {\fontsize{8}{9.5}\selectfont\bfseries\color{LabSlate} Operational business knowledge}\\[1.2pt]
        {\fontsize{6.5}{8}\selectfont {\bfseries\color{LabPurple} Declarative:}} {\fontsize{6.5}{8}\selectfont Data models, business objects \& rules \& process models, ...}\\[1.2pt]
        {\fontsize{6.5}{8}\selectfont {\bfseries\color{LabOrange} Procedural:}} {\fontsize{6.5}{8}\selectfont Agent logic in natural language, application logic as code, ...}
    };

    \node[above=0.35cm of operational.north west, anchor=south west, align=left, text width=5.5cm] (world_text) {
        {\fontsize{9}{10.5}\selectfont\bfseries \textcolor{LabBlueBorder}{(Operational)} \textcolor{LabSlate}{World knowledge}}\\[1.2pt]
        {\fontsize{6.5}{8}\selectfont\color{LabSlate!85} General world knowledge about relevant entities, types, and events}\\[1.2pt]
        {\fontsize{6.5}{8}\selectfont\color{LabBlueBorder} \textbf{Business concepts \& functions, implicit background assumptions and relations}}
    };

    \begin{scope}[on background layer]
        
        \coordinate (left_phantom_edge) at ([xshift=-3.8cm]operational.west);
        \node[green_container, fit=(relational)(operational)(left_phantom_edge), inner sep=5pt] (green_box) {};

        \node[blue_container, fit=(operational)(world_text), inner sep=4pt] (blue_box) {};
        
    \end{scope}

    \node[anchor=center, align=center, font=\bfseries\color{LabSlate}] 
        at ($(green_box.west)!0.5!(green_box.west -| operational.west)$) 
        {
            \fontsize{13}{15}\selectfont Semantically\\[0.5pt]
            \fontsize{13}{15}\selectfont Linked\\[0.5pt]
            \fontsize{13}{15}\selectfont Tables
        };

    \end{tikzpicture}
    \caption{The composition of SLT.}
    \label{fig:slt_overview}
\end{figure}
\looseness=-1 Others employ multi-table relational methods using graph neural networks~\cite{rdl,relbench} that, while effective at capturing relational structures, often assume information completeness within tables, neglecting the operational context essential for understanding data generated by real-world applications. Although current models display impressive generalization capabilities, this oversimplification of tabular data presents a gap that must be addressed. As Sutton's ``bitter lesson'' teaches us~\cite{sutton2019bitter}, the most effective AI progress comes from leveraging computation with general methods rather than hand-crafting domain knowledge. Yet in the case of tabular data, we face a paradox: the operational logic governing data creation is \textit{already written down} in code, constraints, and documentation. Rather than forcing models to rediscover this logic through statistical learning, FMSLTs propose to make it directly accessible, combining the bitter lesson's emphasis on scalable learning with the practical recognition that explicit operational knowledge exists and should be utilized. This integration is visualized in Fig.~\ref{fig:slt_overview}, which illustrates how SLT combines three essential layers: (1) relational data (database tables linked by foreign keys), (2) operational business knowledge (both declarative: data models, business objects \& rules; and procedural: application logic as code, agent logic), and (3) world knowledge (general understanding of business concepts and implicit assumptions). Together, these layers form the ``semantic frame'' that FMSLTs leverage to ground tabular data in its real-world context.
\looseness=-1 Recent work suggests foundation models contain implicit world models~\cite{abdou-etal-2021-language,li2023emergent,vafa2024evaluating}, and the database community is exploring Foundation Database Models~\cite{FoundationDatabaseModel2025}. We hypothesize that for structured data, explicitly modeling semantic context promotes greater generalizability. Position papers on world models~\cite{DingWorldModelSurvey2025,ravi2025world} emphasize counterfactual reasoning and causal understanding, capabilities requiring operational grounding. This motivates \textbf{Semantically Linked Tables (SLT)}, which leverage relational structure alongside semantic relationships for world modeling in structured data. FMSLTs integrate relational data with operational knowledge, constituting a \textbf{new model class}. Tables in real-world systems are relationally linked through database constraints and operational context (business rules, validation code) that governs data operations (Fig.~\ref{fig:slt_overview}).

Standard RAG is insufficient because it retrieves code as text, whereas FMSLTs must ingest code as logic---learning the causal executable semantics (e.g., branching paths) rather than just surface-level text patterns. Consider a concrete example: an expense approval system with the rule \texttt{if amount >= 5000: require\_manager\_approval()}. A statistical model trained on historical data might learn an approximate threshold around \$4,800 from observed correlations, occasionally misclassifying edge cases. A text-based RAG system might retrieve the code snippet but fail to recognize that ``\texttt{>=}'' creates a hard decision boundary. In contrast, an FMSLT ingests this rule as executable logic, understanding that \$4,999 requires no approval while \$5,000 does---achieving perfect accuracy on this policy without any training examples. This distinction between \textit{code-as-text} versus \textit{code-as-logic} is fundamental: models must learn that conditional statements create decision boundaries, loops define iteration semantics, and function calls invoke specific behaviors---not merely that certain tokens co-occur. Recent work on code world models~\cite{metaCWM2025,lehrach2025codeworldmodels} demonstrates that training on execution traces rather than static code significantly improves reasoning capabilities.

\textbf{We propose Foundation Models for Semantically Linked Tables (FMSLT) to integrate this operational knowledge---connecting tables to the logic and rules that govern their creation and evolution.} This grounding encompasses intra- and inter-table relationships, rich contextual metadata, and procedural logic. \textbf{Such grounding is particularly critical for autonomous agents in real-world applications.} Recent enterprise agent benchmarks expose fundamental gaps: WorkArena++~\cite{workarenaplus2024} reveals agents struggle with contextual understanding in knowledge work, TheAgentCompany~\cite{xu2025theagentcompany} shows agents complete only 30\% of professional tasks requiring implicit business knowledge, CRMArena-Pro~\cite{huang-etal-2025-crmarena-pro} documents accuracy drops in multi-turn business dialogues, AgentArch~\cite{agentarch2025} demonstrates agents achieve only 35.3\% success on complex enterprise tasks, MLGym~\cite{mlgym2025} shows frontier models fail to generate novel hypotheses in AI research tasks, and COMPASS~\cite{compass2026} demonstrates that models refuse only 13--40\% of organizational policy violations. FMSLTs address these shortcomings in two ways. First, FMSLTs serve as an \textit{operational world model} grounding agents that access structured knowledge. \textbf{Critically, FMSLTs differ from Text-to-SQL agents}: NL2SQL answers queries (read-only), while FMSLTs \textit{predict state changes} and simulate outcomes---enabling counterfactual reasoning that query engines cannot support. FMSLTs complement autonomous tool-use~\cite{patil2024gorilla,schick2023toolformer}, zero-shot planning~\cite{huang2022language}, and NL2SQL parsing~\cite{li2023can,dail_sql,papicchio2025think2sqlreinforcellmreasoning,biswal2024text2sqlenoughunifyingai} with executable understanding of business logic. Second, FMSLTs enable \textit{AutoML} capabilities, replacing manual feature engineering and model selection as well as current ``data science agents''~\cite{ou2025automind,anonymous2025scaling} that rely on LLMs alone. In both cases, FMSLTs provide the semantic layer for robust autonomy by allowing agents to ``read'' business logic from code and rules rather than guessing from correlations.

\begin{figure*}[t]
    \centering
    \begin{minipage}[b]{0.55\textwidth}
        \centering
        \includegraphics[width=\linewidth]{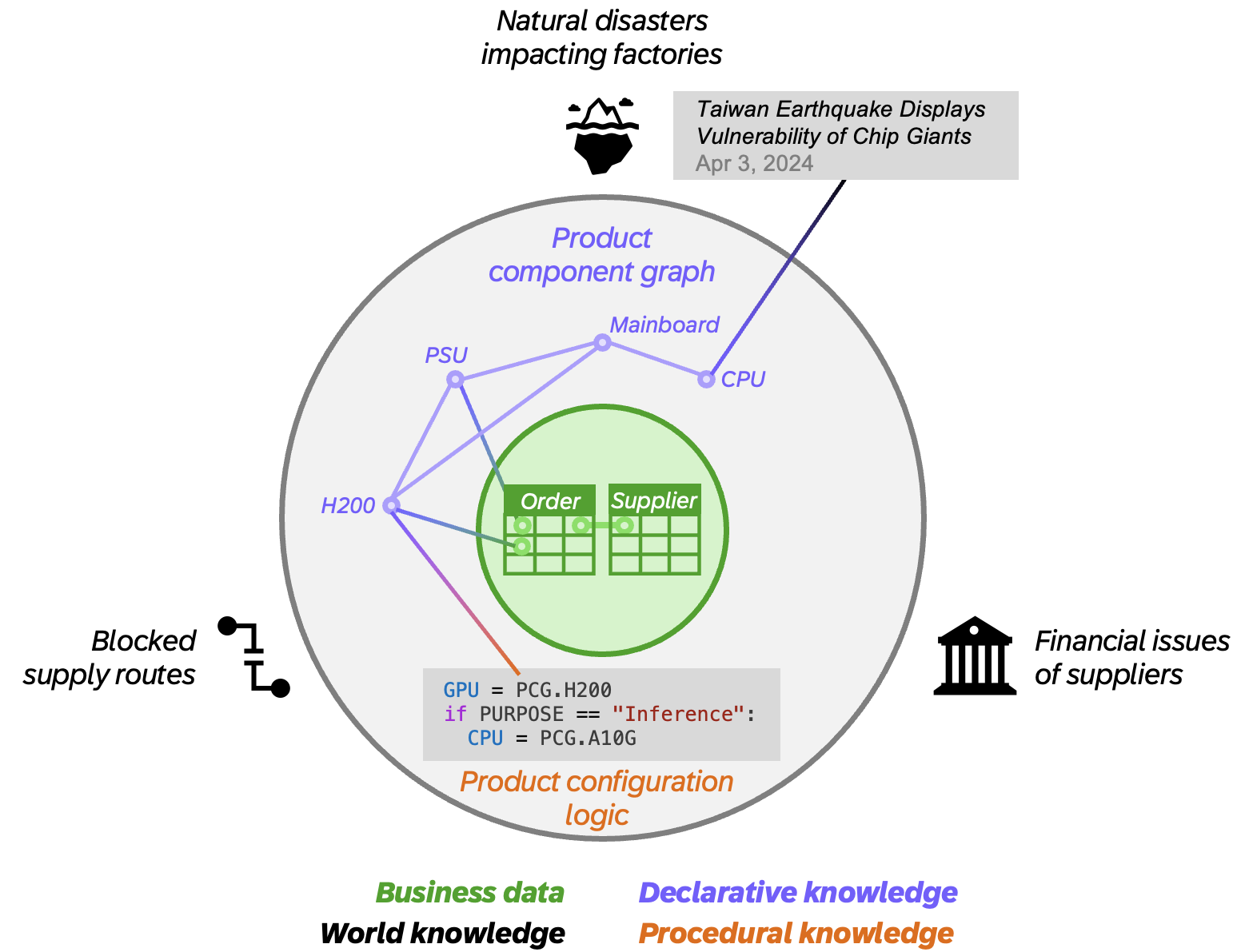}
        \caption{\protect\textbf{SLT Example:} A manufacturing supply chain with webshop integration. Tables are linked through foreign keys and governed by operational logic (configuration rules, inventory validation) that defines valid data states.}
        \label{fig:slt_ex}
    \end{minipage}
    \hfill
    \begin{minipage}[b]{0.41\textwidth}
        \centering
        \includegraphics[width=\linewidth]{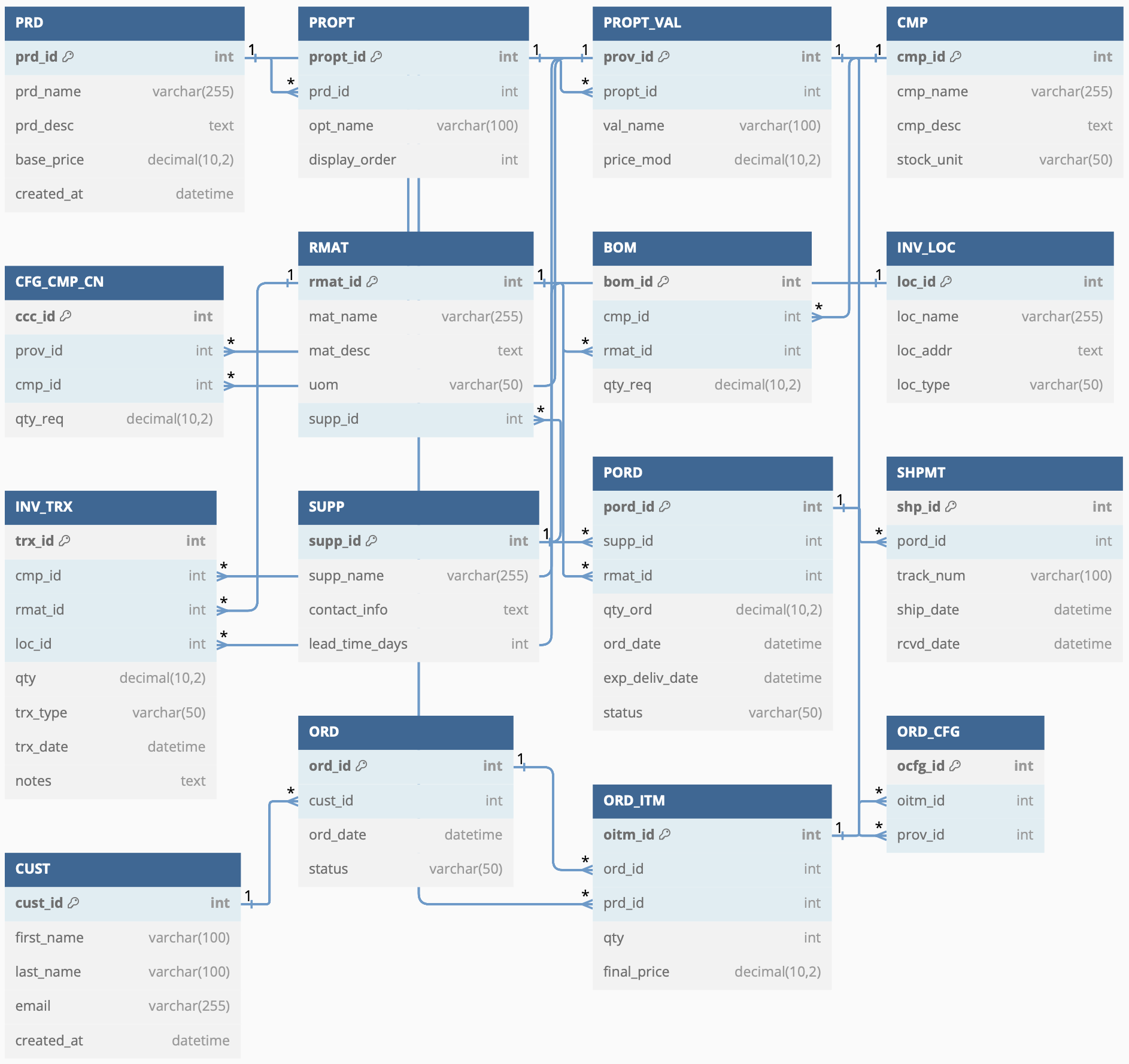}
        \caption{Mockup schema.}
        \label{fig:manu_tab}
    \end{minipage}
\end{figure*}

To illustrate how existing approaches fall short, consider the following example contrasting a vanilla ML approach and an FMSLT within an SLT scenario. Figure~\ref{fig:slt_ex} showcases a simplified supply chain involving a manufacturer of configurable goods (in this case computers) with an associated webshop. The webshop allows the configuration of computers with compatible hardware elements, taking into account information from the warehouse and the availability of items.

Here, SLT encompasses components such as product catalog and products with their configurable components, including warehouse management and supply tracking. For instance, when predicting internal material restocking requirements during production, a typical machine learning approach would be constrained to a company's order history, or perhaps a manually curated subset of data from past analysis, limited by the underlying data complexity c.f.~\cite{githubReleaseAdventureWorks,klein2024salt}. However, effective material restocking requires understanding not only the relational schema (see Fig.~\ref{fig:manu_tab}) but also the operational context---such as the product configuration rules implemented in the web shop's application code---that governs which component combinations are valid and how inventory flows through the system. While this example is drawn from a business context, similar complexities arise in other domains such as healthcare, where operational contexts are equally critical for robust data-driven decision making.

To advance FMSLT capabilities, we address three core requirements: data acquisition, functional architecture, and evaluation methodology. \textbf{Our contributions are threefold:} (1) we define SLT and FMSLT as a new model class; (2) we identify data mixtures necessary for learning grounded representations; and (3) we propose a research roadmap with targeted evaluation benchmarks.

The paper proceeds as follows: Section~2 defines SLT and FMSLT; Section~3 examines data challenges across declarative, procedural, and tabular modalities; Section~4 outlines the FMSLT research roadmap; Section~5 positions FMSLTs relative to existing paradigms; and Section~6 proposes the Operational Turing Test as a benchmark and calls for community collaboration. We recognize that operational knowledge—the procedural logic and business rules governing real-world data—is often proprietary and unavailable in public datasets. Realizing FMSLTs thus requires collaboration between domain experts and ML researchers, complemented by synthetic data generation to simulate operational scenarios while preserving privacy. This position paper articulates the current limitations of tabular foundation models and charts a path toward FMSLTs that achieve robust, context-aware understanding of structured data in its operational context.

\section{Semantically Linked Tables}
\label{sec:slt}
In many real-world environments, the available data resembles an archipelago of semi-isolated information islands. These ``islands,'' which are individual tables, are typically understood only by their creators and domain specialists. Even when organized in relational databases with explicit foreign key relationships connecting tables, each table or cluster of related tables embodies both application-specific details and an implicit conceptualization of the domain. Ideally, table schemas (including names and column headers) would possess semantic richness, enabling direct interpretation. However, in practice, these schema elements frequently act as opaque shorthands, demanding significant additional contextualization~\cite{zhang-etal-2023-nameguess}. This fragmentation into semi-isolated tables substantially hinders the derivation of comprehensive insights and the establishment of meaningful connections across the overall data landscape. This opacity exceeds the limited context offered by relational database systems (e.g., column/table names, foreign keys), which fail to capture the rich operational context of tables. Research on training machine learning models on relational data highlights the challenges of extracting meaning from such interconnected datasets~\cite{relbench}. Insufficient metadata contributes to data ``swamps''~\cite{DBLP:conf/semweb/Mihindukulasooriya23}, where data volume, coupled with semantic ambiguity, hinders information retrieval.

Current efforts focus on metadata enrichment using LLMs to generate metadata and improve concept matching via ontologies~\cite{DBLP:conf/semweb/Mihindukulasooriya23}, enhancing dataset search~\cite{Brickley2019GoogleDS,anonymous2024tabmeta} and downstream tasks like forecasting~\cite{williams2025contextkeybenchmarkforecasting} and analysis~\cite{turl}.

\begin{tcolorbox}[colback=gray!5, colframe=gray!40, boxrule=0.5pt, arc=2pt, left=6pt, right=6pt, top=6pt, bottom=6pt]
\begin{definition}[Operational Knowledge]
\label{def:operational_knowledge}
\textbf{Declarative Knowledge} (the ``What''): Domain concepts, rules, and constraints that define valid states and relationships---typically encoded in knowledge graphs, ontologies, policy documents, and configuration files.
\textbf{Procedural Knowledge} (the ``How''): Executable logic governing data creation and manipulation---encoded in source code, validation scripts, workflows, and application logic that dictates how systems update state.
\end{definition}
\end{tcolorbox}

\subsection{Elements of Foundation Models for SLT}
\looseness=-1 FMSLTs ground relational data by integrating both forms of operational knowledge (Definition~\ref{def:operational_knowledge}), requiring curated data mixtures~\cite{olmo20252olmo2furious} to provide the ``missing link'' to real-world context:

\looseness=-1 \noindent\textbf{1. Declarative Knowledge:} In our webshop example (Fig.~\ref{fig:slt_ex}), this includes compatibility policies (``Mainboard X supports only DDR5'') and product catalogs. While traditionally captured in symbolic knowledge graphs (KGs)~\cite{Gruber1995TowardPF}---from general ones like Wikidata~\cite{vrandevcic2014wikidata} to corporate implementations~\cite{hogan2025largelanguagemodelsknowledge}---connecting logical rigidity with neural adaptability remains a neurosymbolic challenge~\cite{marra2024statisticalrelationalneurosymbolicartificial}. FMSLTs must leverage recent advances in CoT~\cite{NEURIPS2022_9d560961} (despite linguistic limits~\cite{MAHOWALD2024517}) and graph foundation models~\cite{Mao2024PositionGF} to ingest these explicit rules, ensuring models adhere to compliance constraints.

\looseness=-1 \noindent\textbf{2. Procedural Knowledge:} This includes the code validating a checkout selection or computing inventory requirements. Unlike static schemas, procedural knowledge captures operation semantics. Since LLMs already excel at code generation~\cite{Chen2021EvaluatingLL} (e.g., Code Llama~\cite{roziere2024}, DeepSeek Coder~\cite{Guo2024DeepSeekCoderWT}), FMSLTs can naturally leverage source code repositories to learn this underlying functionality, bridging the gap between static data observations and the dynamic processes that generated them.

\section{Data Challenges}
\label{sec:data_challenges}
\looseness=-1 Data silos~\cite{bruckhaus2024ragdoesworkenterprises,Urlana2024LLMsWI} remain a primary obstacle to realizing AI's full potential, particularly in machine learning and deep learning. These silos are prevalent across different domains, exhibiting considerable similarities, especially in complex settings like healthcare and business operations. Both types of applications face challenges in data governance and access restrictions, stemming from disparate systems governed by varying policies, security protocols, and access controls. In business applications, fragmentation arises from competitive sensitivities, departmental divisions, or mergers and acquisitions. Healthcare is constrained by stringent privacy regulations (e.g., HIPAA, GDPR), patient consent requirements, and institutional policies~\cite{Luc2019}. Furthermore, both domains struggle with data heterogeneity and standardization, including inconsistent terminologies, varying data quality, and structural variations. Healthcare's issues are compounded by variations in data acquisition protocols and equipment manufacturers~\cite{Rieke2020TheFO}, while complex operational systems often involve intricate knowledge bases spanning multiple domains, formats, and systems~\cite{journals/corr/abs-2310-11703}. These shared issues impede AI development by limiting the creation of large and diverse datasets essential for effective deep learning~\cite{vanPanhuis2014ASR,8237359,doi:10.1148/rg.2017170077}. A key objective is to circumvent these data siloing pitfalls and promote a more integrated approach to data utilization, potentially using synthetic data generation and privacy-enhancing techniques.

\subsection{Declarative Data}
\looseness=-1 Large, heterogeneous declarative data assets are increasingly managed as knowledge graphs (KGs) by major organizations~\cite{10.1145/3331166}, illustrated by Google's Knowledge Graph, Microsoft Academic Graph, and Bloomberg's Financial KG, with broad industry adoption documented by~\cite{hogan2021knowledge, Tamasauskaite2023DefiningAK}. These KGs, exemplified by the recent SALT-KG benchmark~\cite{SALTKG2025}, often contain proprietary, confidential, and sensitive information. KG grounding has been shown to improve factuality and reduce hallucinations~\cite{guan2024hallucination,luo2024factuality,sun2024pog}, with empirical studies demonstrating substantial improvements in reasoning accuracy when structured knowledge graphs are integrated with LLMs compared to unstructured text alone. However, creating systems that can learn from these sensitive assets to enable transferability and inductive reasoning across different organizations remains a key objective.
\looseness=-1 Sharing or synthesizing KGs without violating privacy is an extremely challenging problem. Open-sourcing even partial KGs is generally infeasible, and generating synthetic data that preserves both privacy and semantic utility remains a significant technical challenge. Furthermore, public and domain-specific KGs often comprise completely disjoint entity and relation sets, making it difficult to train universally transferable models.
\looseness=-1 To address these and related challenges, recent research investigates the development of foundational models for (knowledge) graphs (FMGs)~\cite{galkin2024towards,galkin2024ultraquery}. FMGs are designed to learn universal and transferable graph representations, enabling inference over unseen nodes and relations. By adopting inductive generalization properties, these models can facilitate reasoning across diverse graphs with differing vocabularies. Training FMGs on open-source KGs represents an initial step toward this vision, but there remains a critical need for developing privacy-preserving capabilities to enable secure collaboration and deployment in real-world scenarios.

\subsection{Procedural Data}
\looseness=-1 While declarative knowledge defines \textit{what} the rules are, procedural knowledge captures \textit{how} systems execute them. Unlike KGs which excel at representing static entities and relationships, procedural knowledge requires capturing dynamic execution semantics. Current coding LLMs train primarily on public corpora (GitHub, Stack Overflow), which often lack the proprietary operational logic governing AI systems. This gap is critical for SLT, where procedural components (state transitions from application logic, validation scripts, workflows) are essential for understanding how declarative rules manifest as executable operations on tables (see Appendix~\ref{appendix:supply_chain} for a concrete validation algorithm an FMSLT must internalize). \textbf{FMSLTs extend beyond declarative KGs by treating procedural code as a first-class citizen}: directly learning the dynamic ``verbs'' (e.g., execution semantics of \texttt{check\_inventory()}) that KGs cannot structurally represent. Program-of-Thought approaches~\cite{khatuya2025finder} demonstrate that generating intermediate code substantially improves reasoning on financial/tabular data, reinforcing the need for FMSLTs to internalize rather than merely retrieve procedural logic. Integrating proprietary procedure repositories could deepen FMSLTs' understanding of \textit{Code-to-Data} relationships.

\noindent\textbf{Synthetic code generation:}
Synthetic code generation offers a privacy-preserving solution. Recent work shows synthetic data substantially improves code models~\cite{hui2024qwen25codertechnicalreport}, particularly through agentic frameworks combining symbolic guidance with neural generation~\cite{ni2024treeofcodetreestructuredexploringframework,openhands}. Crucially, code models trained on execution traces---observing code-data interactions at runtime---learn superior world models versus static code alone~\cite{metaCWM2025}. This supports the FMSLT hypothesis: grounding in operational dynamics (code execution paired with table states) yields better reasoning than pure syntactic patterns. By leveraging these approaches, FMSLTs could generate synthetic training examples consisting of procedural code (e.g., validation functions, workflow logic) paired with corresponding table data and execution traces, enabling privacy-preserving pre-training on realistic operational behaviors without exposing proprietary systems.

\subsection{Tabular Data}
\looseness=-1 Most tabular research relies on web-scraped ``information islands'' (e.g., WebTables~\cite{10.14778/1453856.1453916}, TabLib~\cite{eggert2023tablib}), missing operational context. While datasets like TURL~\cite{turl} or GitTables~\cite{hulsebos2023gittables} are cleaner, they lack interconnectedness. Recent datasets like SQaLe~\cite{wolff2025sqale}, RelBench~\cite{pmlr-v235-fey24a}, and Adventure Works~\cite{githubReleaseAdventureWorks} introduce multi-table scenarios with relational interconnectedness, but still lack the full operational knowledge of SLT. This gap emphasizes the need for datasets grounded in real-world complexity to advance FMSLT.

\noindent\textbf{Synthetic tables:}
\looseness=-1 Deep generative models (GANs, VAEs) create privacy-preserving synthetic tables~\cite{xu2019modeling}. Recent work extends this to cross-tabular generation~\cite{CTSyn2025}, yet capturing complex inter-table logic remains challenging, though recent diffusion-based approaches~\cite{pang2024clavaddpm,ketata2025grdm} show promise. LLM-based generation~\cite{borisov2023language} offers potential but struggles with strict schema adherence and scale. Recent work on Real-TabPFN~\cite{garg2025realtabpfn} highlights the value of real-world tabular data for foundation models, reinforcing our call for comprehensive, grounded data strategies.

Tabular data is the product of both declarative and procedural processes, intrinsically merging these two distinct knowledge sources. This dual nature presents both opportunities and challenges: maintaining operational integrity while ensuring sufficient diversity becomes complex. Privacy concerns restrict access to real-world operational data, making synthetic data generation vital for creating realistic yet anonymized datasets. Such datasets~\cite{klein2024salt} are crucial for advancing FMSLT research. Single-table synthesis has progressed significantly~\cite{pmlr-v202-kotelnikov23a,shi2024tabdiffmultimodaldiffusionmodel,kwok2025greatergeneraterealistictabular}, but SLT requires multi-table synthesis, a more complex challenge. Existing approaches (Synthetic Data Vault~\cite{SDV}, PrivLava~\cite{10.1145/3589287}) face scalability limitations. Diffusion-based methods~\cite{pang2024clavaddpm,ketata2025grdm,rombach2022high} show promise for joint relational database generation. Beyond generative models, simulators~\cite{chang2024learningproductionfunctionssupply} and digital twins~\cite{Tornqvist2024ATA,wang2024twingptdigitaltwinsclinical} offer viable approaches, with secure sandboxes enabling privacy-preserving benchmarking~\cite{rajore2024truceprivatebenchmarkingprevent}.

\subsection{Breaking the Privacy Deadlock}
\looseness=-1 The circularity of needing private data to train synthetic generators is resolvable via two-stage bootstrapping. First, FMSLTs pre-train on open-source ecosystems (GitHub repos with SQL databases) and simulator-generated synthetic data, learning ``Code-to-Data'' relationships in a privacy-neutral setting. \textbf{Crucially, generators must avoid ``too clean'' data} by injecting noise, version conflicts, and ``simulated technical debt'' (e.g., legacy data violating current rules). This prevents overfitting to ``perfect logic,'' preparing models for real-world messiness. This \textit{Curriculum Learning} has two phases: \textbf{Phase 1} teaches the \textit{meta-skill} of reading code (``Causal Mechanics''); \textbf{Phase 2} fine-tunes this for specific domains. Open-source ERPs and public data provide rich examples without exposing secrets. This teaches how procedural code governs data---a transferable skill. Second, federated learning~\cite{10.1145/3589287} or differential privacy enables adaptation to private data without centralization. This ``bootstrapping from openness'' learns operational semantics publicly, then specializes privately.

\section{The FMSLT Framework}
\label{sec:towards}
We outline the landscape of tabular architectures and the future of operational world models.

\looseness=-1 \noindent\textbf{Neural Tabular Models:}
Neural models for tabular data are advancing rapidly~\cite{somvanshi2024surveydeeptabularlearning,Fang2024,badaro-etal-2023-transformers}, challenging tree-based dominance~\cite{grinsztajn2022why,Chen:2016:XST:2939672.2939785,Catboost2021,ye2025closerlookdeeplearning,holzmueller2024better}. However, current models treat tables as isolated entities, neglecting operational context despite tabular data's distinct challenges~\cite{gardner2024large,raman2024scalablerepresentationlearningmultimodal,hu2024pytorch,spinaci2024portal,su2024tablegpt2largemultimodalmodel}.
\looseness=-1 LLMs struggle with temporal~\cite{islakoglu2025chronosenseexploringtemporalunderstanding} and causal~\cite{nastl2024do} reasoning in this domain. Recent ICL approaches (TabPFN-2.5~\cite{TabPFN25}, ConTextTab~\cite{ConTextTab2025}, RealMLP~\cite{holzmueller2024better}) now match gradient boosting~\cite{TabArena2025,hollmann2025nature}, though early work focused on small-scale data~\cite{Binder,10.1145/3539618.3591708,ChainOfTable,hollmann2025nature}. While demonstrating strong single-table capabilities, they do not yet incorporate the operational context central to SLT. The lack of large, standardized datasets capturing multi-table operational relationships further hinders progress toward fully grounded tabular foundation models.
\looseness=-1 Scalability is advancing through deeper architectures~\cite{ma2024tabdptscalingtabularfoundation,hollmann2024tabpfn,su2024tablegpt2largemultimodalmodel} and improved in-context learning~\cite{qu2025tabicl,TabDPT2025}, with document-based contextualization enabling competitive zero-shot performance~\cite{wydmuch2024tacklingpredictiontasksrelational}.
Graph-like representations offer a promising direction for Foundation Models for Graphs (FMGs). \cite{CARTE} employs star-shaped graphlets and graph-attentional networks to contextualize table entries. Similarly, the Zero-Shot Relational Transformer~\cite{RT2025} demonstrates the potential of graph-based architectures for relational reasoning. However, while capturing static structure, such methods lack the ability to model \textit{dynamic} data evolution governed by procedural rules. This is a gap FMSLT aims to bridge.

\looseness=-1 \noindent\textbf{Operational World Model:}
Integrating world models is critical for FMSLT to mirror real-world dynamism. Recent work shows LLMs can serve as implicit world models~\cite{liu2024wordtoworld}, yet effectiveness depends on behavioral coverage~\cite{wang2024llmworldmodels}. Executable code-based world models outperform direct LLM inference~\cite{lehrach2025codeworldmodels}. \textbf{A Digital Twin is a simulation; an FMSLT is a \textit{simulator generator}.} When rules change, a Digital Twin must be re-coded by engineers; an FMSLT simply reads the updated policy PDF or source code and adapts zero-shot. This zero-shot adaptability (generalizing to new operational logic without re-engineering) is the critical AI advantage over hard-coded simulations. Future work should unify these paradigms using object-centric methods~\cite{pmlr-v162-zhao22b}.

\looseness=-1 \noindent\textbf{Symbolic Representation of World Knowledge:}
A critical distinctness of FMSLT is its requirement to ingest and reason over symbolic artifacts (documents, code, and constraints), not merely as text, but as executable or constraining logic. Points raised in recent discussions highlight that declarative and world knowledge can be symbolically represented, offering a rigorous scaffolding for learning. Neuro-symbolic approaches offer a promising path here, allowing models to learn differentiable representations of symbolic rules. For example, a declarative business rule (e.g., ``approvals required for expenses over $\ge$\textdollar 5k'') should not act solely as a statistical correlation but as a logical constraint in the model's inference process. Emerging work in neuro-symbolic AI~\cite{nye2021improving,d2024neurosymbolic,yang2025neurosymbolicsurvey} suggests that hybrid architectures can effectively bridge this gap. \textbf{Unlike traditional neuro-symbolic systems requiring rigid logic definitions}, FMSLTs learn \textit{soft consistency}, internalizing business logic ``physics'' through pre-training. Crucially, FMSLTs don't just \textit{execute} rules; they \textit{predict the outcome} of rule execution, including cases where rules are overridden (e.g., an executive bypass) or inconsistently applied. Chain of Code~\cite{li2024chainofcode} introduces the ``LMulator'' concept (using an LLM to simulate code execution when symbolic execution is infeasible), which supports our argument that FMSLTs can reason over operational artifacts even when they are not directly runnable (e.g., policy documents). This ``probabilistic execution'' capability (predicting what \textit{will} happen rather than what \textit{should} happen) distinguishes FMSLTs from rigid symbolic interpreters.

\looseness=-1 \noindent\textbf{Algorithmic Alignment via Synthetic ICL:}
We propose that FMSLTs should be trained primarily on synthetic ``System-Table'' pairs. By procedurally generating business logic (e.g., Python validation scripts) and corresponding synthetic table states, we can train the model on an infinite stream of ``Reasoning Tasks.'' This shifts the learning objective: instead of memorizing the correlations of a static dataset (which risks privacy leakage), the model learns the meta-capability of In-Context Learning (ICL), applying an arbitrary, unseen rule to a new table row. This effectively bypasses the privacy deadlock, as the model learns \textit{how to reason} from synthetic data, but applies that reasoning to private data only at inference time.

\begin{figure*}[t]
\centering
\begin{minipage}[t]{0.5\textwidth}
\centering
\begin{tikzpicture}[
    scale=1.1, transform shape,
    every node/.style={font=\sffamily\tiny},
    pill/.style={
        draw,
        rectangle,
        rounded corners=0.2cm,
        minimum height=0.75cm,
        align=center,
        line width=0.8pt,
        drop shadow={opacity=0.25, shadow xshift=0.5pt, shadow yshift=-0.5pt}
    },
    purple pill/.style={pill, fill=violet!10, draw=violet!80!black, text=violet!60!black, minimum width=2.0cm},
    red pill/.style={pill, fill=red!10, draw=red!80!black, text=red!60!black, minimum width=2.0cm},
    blue pill/.style={pill, fill=blue!10, draw=blue!60!black, text=blue!40!black, minimum width=1.5cm},
    green pill/.style={pill, fill=teal!10, draw=teal!80!black, text=teal!60!black, minimum width=1.5cm},
    arrow/.style={-Stealth, thick, draw=black!75, shorten >=2pt, shorten <=1pt},
    dashed arrow/.style={-Stealth, thick, draw=violet!80!black, dashed, line width=0.8pt, shorten >=2pt, shorten <=1pt}
]

\node[purple pill] (syn_rules) at (1.1, 1.8) {\textbf{Synthetic}\\[-1pt]\textbf{Rules}};
\node[purple pill] (syn_data) at (1.1, 0.8) {\textbf{Synthetic}\\[-1pt]\textbf{Tables}};
\node[blue pill] (learn) at (3.8, 1.3) {\textbf{Learn}\\[-1pt]$P(\text{Row}|\text{Rule})$};
\node[green pill] (icl) at (6.0, 1.3) {\textbf{ICL}\\[-1pt]\textbf{Skill}};

\draw[arrow] (syn_rules.east) -- (learn.west);
\draw[arrow] (syn_data.east) -- (learn.west);
\draw[arrow] (learn.east) -- (icl.west);

\node[red pill] (priv_docs) at (1.1, -0.6) {\textbf{Declarative}\\[-1pt]\textbf{Knowledge}};
\node[red pill] (priv_code) at (1.1, -1.6) {\textbf{Procedural}\\[-1pt]\textbf{Knowledge}};
\node[red pill] (priv_db) at (1.1, -2.6) {\textbf{Relational}\\[-1pt]\textbf{Data}};
\node[blue pill] (retrieval) at (3.8, -1.6) {\textbf{Retrieval +}\\[-1pt]\textbf{Zero-Shot}};
\node[green pill] (prediction) at (6.0, -1.6) {\textbf{Prediction}};

\draw[arrow] (priv_docs.east) -- (retrieval.west);
\draw[arrow] (priv_code.east) -- (retrieval.west);
\draw[arrow] (priv_db.east) -- (retrieval.west);
\draw[arrow] (retrieval.east) -- (prediction.west);

\draw[dashed arrow] (icl.south) -- (retrieval.north) node[midway, anchor=west, font=\sffamily\tiny, text=black!70, align=left, xshift=6pt] {Transfer\\Meta-Skill};

\begin{scope}[on background layer]
    \node[
        fill=violet!5,
        draw=violet!20,
        rounded corners=5pt,
        inner sep=6pt,
        fit=(syn_rules) (syn_data) (learn) (icl),
        label={[anchor=north east, font=\sffamily\tiny\bfseries, text=violet!60!black, xshift=-2pt, yshift=-3pt]north east:Phase 1: Synthetic Training}
    ] (train_bg) {};

    \node[
        fill=red!5,
        draw=red!20,
        rounded corners=5pt,
        inner sep=6pt,
        fit=(priv_docs) (priv_code) (priv_db) (retrieval) (prediction),
        label={[anchor=south east, font=\sffamily\tiny\bfseries, text=red!60!black, xshift=-2pt, yshift=2pt]south east:Phase 2: Private Inference}
    ] (inf_bg) {};
\end{scope}

\end{tikzpicture}
\caption{\protect\textbf{Retrieval-Augmented In-Context Learning Architecture.} Training learns reasoning from synthetic system-table pairs; inference retrieves operational artifacts and applies learned ICL for zero-shot prediction on private tables.}
\label{fig:architectural_sketch}
\end{minipage}
\hfill
\begin{minipage}[t]{0.48\textwidth}
\centering
    \begin{tikzpicture}[scale=0.6, font=\sffamily]
        \small
        
        \definecolor{accentblue}{RGB}{59, 130, 246}      
        \definecolor{accentpurple}{RGB}{139, 92, 246}    
        \definecolor{accentgreen}{RGB}{16, 185, 129}     
        \definecolor{bgstart}{RGB}{248, 250, 252}        
        \definecolor{bgend}{RGB}{241, 245, 249}          
        \definecolor{gridcolor}{RGB}{203, 213, 225}      
        \definecolor{textgray}{RGB}{71, 85, 105}         
        \definecolor{baseline}{RGB}{148, 163, 184}       
        
        \shade[top color=bgstart, bottom color=bgend] (0, 0) rectangle (9.0, 4.7);
        
        \draw[gridcolor, very thin, opacity=0.3] (0, 1.5) -- (9.0, 1.5);
        \draw[gridcolor, very thin, opacity=0.3] (0, 3.0) -- (9.0, 3.0);
        \draw[gridcolor, very thin, opacity=0.3] (0, 4.5) -- (9.0, 4.5);
        \draw[gridcolor, very thin, opacity=0.3] (2.2, 0) -- (2.2, 4.7);
        \draw[gridcolor, very thin, opacity=0.3] (4.4, 0) -- (4.4, 4.7);
        \draw[gridcolor, very thin, opacity=0.3] (6.6, 0) -- (6.6, 4.7);
        
        \draw[textgray, thick, ->] (0, 0) -- (0, 5.0);
        \draw[textgray, thick, ->] (0, 0) -- (9.2, 0);
        
        \draw[textgray, thick] (-0.15, 0.0) -- (0.15, 0.0);
        \node[align=right, font=\small, text=textgray, anchor=east] at (-0.25, 0.5) {Single\\Table};
        
        \draw[textgray, thick] (-0.15, 2.25) -- (0.15, 2.25);  
        
        \draw[textgray, thick] (-0.15, 2.3) -- (0.15, 2.3);
        \node[align=right, font=\small, text=textgray, anchor=east] at (-0.25, 2.3) {Relational\\Data};
        
        \draw[textgray, thick] (-0.15, 4.5) -- (0.15, 4.5);
        \node[align=right, font=\small, text=textgray, anchor=east] at (-0.25, 4.5) {Semantically\\Linked Data};
        
        \draw[textgray, thick] (0.0, -0.15) -- (0.0, 0.15);
        \node[align=center, font=\small, text=textgray, anchor=north] at (0.0, -0.25) {Single-Task\\Model};
        
        \draw[textgray, thick] (3.3, -0.15) -- (3.3, 0.15);
        \node[align=center, font=\small, text=textgray, anchor=north] at (3.3, -0.25) {Multi-Task\\Model};
        
        \draw[textgray, thick] (6.6, -0.15) -- (6.6, 0.15);
        \node[align=center, font=\small, text=textgray, anchor=north] at (6.6, -0.25) {Foundation\\Model};
        
        \node[circle, draw=baseline, fill=baseline!20, line width=1.2pt, inner sep=3.5pt] (gbdt) at (0.6, 0.4) {};
        \node[font=\small, text=textgray, anchor=west] at (0.85, 0.4) {GBDTs};
        
        \node[circle, draw=baseline, fill=baseline!20, line width=1.2pt, inner sep=3.5pt] (relben) at (0.6, 2.25) {};
        \node[font=\small, text=textgray, anchor=west] at (0.85, 2.25) {RelBen};
        
        \node[circle, draw=baseline, fill=baseline!20, line width=1.2pt, inner sep=3.5pt] (fmdb) at (3.3, 2.25) {};
        \node[font=\small, text=textgray, anchor=west] at (3.55, 2.25) {FMDB};
        
        \node[circle, draw=baseline, fill=baseline!20, line width=1.2pt, inner sep=3.5pt] (ltm) at (6.6, 0.4) {};
        \node[font=\small, text=textgray, anchor=west] at (6.85, 0.4) {LTM};
        
        \node[circle, draw=accentblue, fill=accentblue!20, line width=1.2pt, inner sep=3.5pt] (fmg) at (6.6, 3.4) {};
        \node[font=\small, text=textgray, anchor=west] at (6.85, 3.4) {FMG};
        
        \node[circle, draw=accentgreen, fill=accentgreen!25, line width=2pt, inner sep=4.5pt, drop shadow={opacity=0.15, shadow xshift=0.1cm, shadow yshift=-0.1cm}] (fmslt) at (6.6, 4.5) {};
        \node[font=\bfseries\small, text=accentgreen!80!black, anchor=west] at (6.9, 4.5) {\textbf{FMSLT}};
        
    \end{tikzpicture}
\caption{Conceptual positioning of \protect\textit{FMSLT} relative to existing paradigms, categorized by data richness (y-axis) and model capabilities (x-axis).}
\label{fig:altviews}
\end{minipage}
\end{figure*}

\looseness=-1 This \textit{Retrieval-Augmented In-Context Learning} approach (see Figure~\ref{fig:architectural_sketch}) comprises: (1) \textit{Training Phase}: The model learns to predict table row outcomes given arbitrary business rules by training on procedurally-generated synthetic data. This teaches the meta-capability of applying unseen logic to structured data. Crucially, this relies on the hypothesis that logical reasoning is modality-invariant, analogous to \textit{Code-Induced Reasoning} where code pre-training improves general reasoning tasks~\cite{Huang2022LanguageMA,Chen2021EvaluatingLL,lehrach2025codeworldmodels,metaCWM2025}. We posit that by training on diverse synthetic modalities, FMSLTs acquire a generalized `reasoning engine' for zero-shot transfer to operational artifacts. (2) \textit{Inference Phase}: A retriever identifies relevant operational artifacts (declarative knowledge: rules, KGs; procedural knowledge: source code; and relational data) from the private organizational context, and the model applies its learned ICL capability to make zero-shot predictions on private tables. Engineering challenges include designing effective synthetic data generators that capture real-world operational complexity, and building retrievers that can identify semantically relevant artifacts from large organizational systems. \textbf{A critical challenge is the ``Provenance Gap''}: mapping database rows to the specific code functions that generated or validated them. While modern microservices with comprehensive logging provide this traceability, retrofitting provenance for legacy systems remains an open problem. Recent advances in retrieval-augmented generation~\cite{lewis2021retrievalaugmentedgenerationknowledgeintensivenlp} and program synthesis suggest viable paths forward. This architectural sketch demonstrates that FMSLT is not merely a conceptual proposal but a tractable engineering goal requiring innovation beyond current foundation model capabilities.

\enlargethispage{2\baselineskip}
\section{Alternative Views}
\label{sec:altviews}

\textbf{We contend that research on tabular foundation models has pursued a fundamentally narrow path.} Most work focuses on single-table generalization or schema-level relational patterns, ignoring the operational context that defines how data acquires meaning in real systems. Figure~\ref{fig:altviews} organizes existing approaches along two axes: \textit{data richness} and \textit{model capability}. FMSLT represents a necessary paradigm shift to the upper-right quadrant (foundation models on semantically linked data), which remains fundamentally underexplored.

\noindent\textbf{Single-table methods} (GBDTs, TabPFN~\cite{hollmann2025nature,TabPFN25}, ConTextTab~\cite{ConTextTab2025}) occupy the lower portion of Fig.~\ref{fig:altviews}. These excel at isolated table tasks but, by design, neglect relational interconnectedness and operational context. Recent theoretical work reveals fundamental limitations of embedding-based approaches: regardless of dimension, single-vector embeddings cannot scale to combinatorial query complexity~\cite{anonymous2025on}. This mathematical constraint reinforces our position that operational grounding requires explicit integration of procedural knowledge rather than attempting to compress all semantic information into learned representations. The position paper by~\cite{pmlr-v235-van-breugel24a} argues for this scope, and we acknowledge its validity for narrow, static tasks. However, we contend this view is fundamentally limiting for agentic systems where tables lack the semantic self-containment of text~\cite{Saussure1916}.

\noindent\textbf{Relational methods} (RelBench~\cite{relbench}, GraphSAGE~\cite{GraphSAGE}, CARTE~\cite{CARTE}, Griffin~\cite{wang2025griffin}, FMDB, Zero-Shot Relational Transformer~\cite{RT2025}) advance to the middle tier by capturing multi-table dependencies via foreign keys. While valuable, these assume all necessary context is encoded in relational schemas, missing operational knowledge in code and business rules. They also lack the \textbf{In-Context Learning (ICL)} capability of foundation models, requiring retraining for schema changes.

\noindent\textbf{Emerging foundation models} (LTM, FMG~\cite{galkin2024towards}, KumoRFM~\cite{KumoRFM2025}, TabICL~\cite{qu2025tabicl}) extend toward the upper-right but remain within database boundaries. FMSLT completes this trajectory by integrating declarative, procedural, and world knowledge, enabling not just prediction but agentic reasoning: identifying \textit{why} a supply chain delay occurred (new regulation + missing code handler) rather than merely \textit{that} it will occur~\cite{workarenaplus2024,xu2025theagentcompany}.

\noindent\textbf{The world model analogy:} Just as robots need a \textit{physics engine} to predict object motion, business AI requires an \textit{operational model} for data evolution~\cite{li2023emergent,liu2024worldmodelsurvey}. Tables are observations; operational context is the physics engine. Current models infer physics from correlations, akin to rediscovering F=ma by observing falling objects. FMSLTs, by contrast, are \textbf{given} the operational formulas (code, constraints) directly, learning to \textit{apply} them rather than reinvent them from data. This enables counterfactual reasoning (``What if this rule changed?'') that data-driven models cannot achieve. \textbf{We contend that operational grounding is not optional but essential for foundation models that can truly reason, explain, and act.}

\section{Call to Action}
\label{sec:calltoaction}
\textbf{The path forward requires community-wide action.} FMSLTs will not emerge from incremental improvements to existing methods. We call upon the machine learning and data management communities to pursue four concrete, urgent directions:

\noindent\textbf{1. The Operational Turing Test:} \textbf{We contend that current benchmarks fundamentally fail to test whether models truly understand structured data.} They measure statistical generalization, not operational reasoning. We propose a benchmark that no existing tabular foundation model can pass: one that demands genuine comprehension of business logic rather than pattern memorization.

\begin{tcolorbox}[colback=gray!5, colframe=gray!40, boxrule=0.5pt, arc=2pt, left=6pt, right=6pt, top=6pt, bottom=6pt]
\begin{definition}[The Operational Turing Test]
A model passes the Operational Turing Test if it can join a simulated organization (unseen during training) and correctly predict the outcome of a transaction \textit{and explain the decision by citing the relevant operational artifacts} solely by reading the organization's \textbf{Declarative knowledge} (rules and constraints), \textbf{Procedural knowledge} (source code and application logic), and \textbf{Relational data}, with only a \textbf{few-shot} number of labeled input-output examples (appropriate to task complexity) demonstrating the organization's operational behavior.
\end{definition}
\end{tcolorbox}

\textbf{This test is deliberately hard.} Success requires consulting operational artifacts (reading approval logic from code, checking policy PDFs for constraints, traversing foreign key relationships) rather than memorizing statistical patterns. It measures whether a model truly \textit{understands} operational semantics or merely interpolates training data. \textbf{We assert that no existing tabular foundation model can pass this test.} Current benchmarks like RelBench and TabArena test schema generalization on abundant training data, not zero-shot operational reasoning. The Operational Turing Test rectifies this: it demands the ability to make informed predictions about unseen business logic by grounding in explicit operational knowledge, precisely the capability FMSLTs aim to unlock. For example, given a \$6,000 expense submission, a passing model would output: ``Rejected: violates approval threshold defined in \texttt{policy\_v3.pdf} Section 4.2 and enforced in \texttt{approval.py:L47}: \texttt{if amount >= 5000: require\_manager\_approval()}.'' This requires reading both the policy document (declarative) and the source code (procedural) to ground the decision in operational artifacts. We call on industry partners to contribute sanitized operational contexts (anonymized ERP vendors, simulated enterprise environments) and on researchers to develop models capable of passing this benchmark.

\noindent\textbf{2. Develop Privacy-Preserving Data Sandboxes:} \textbf{Privacy is the primary bottleneck to FMSLT progress.} Without access to operational context, models cannot learn grounded reasoning. We insist that industry and academia collaborate on secure execution environments where code and logic can be shared without exposing raw sensitive rows~\cite{rajore2024truceprivatebenchmarkingprevent}. This will enable ``blind' pre-training corpora where models learn procedural dynamics without memorizing PII, leveraging privacy-enhancing techniques such as differential privacy~\cite{10.1145/3589287}.

\noindent\textbf{3. Advance Privacy-Preserving Synthetic Data Generation:} As argued in Section~\ref{sec:data_challenges}, privacy regulations (HIPAA, GDPR) and data silos severely restrict access to the real-world SLT data needed to train FMSLTs. Synthetic data generation is not merely a convenience but a necessity for circumventing these barriers. We call for investment in methods that generate high-fidelity, multi-table synthetic data preserving complex relational dependencies and operational semantics. Recent advances in graph-conditional diffusion models~\cite{ketata2025grdm} and cross-tabular synthesis~\cite{CTSyn2025} demonstrate promising scalability, yet significant research remains to capture procedural constraints, temporal dynamics, and the full semantic richness inherent in SLT---while maintaining rigorous privacy guarantees.

\noindent\textbf{4. Rethink Architectural Paradigms:} Tabular FMs must move beyond retrofitting LLM architectures or scaling up gradient boosting. We call for novel architectures purpose-built for SLT: models that natively process relational structures, code, and rules while maintaining the ICL capabilities of foundation models. Whether via table-native pre-training~\cite{qu2025tabicl} or verbalized structural links~\cite{GTLWen}, FMSLTs demand architectures that unify statistical rigor with semantic understanding.

\section{Conclusion}
\label{sec:conclusion}
We argue that the next breakthrough in tabular AI requires grounding tables in their SLT context---shifting from learning \textit{from} tables to learning \textit{about} the systems that generate them. By integrating declarative, procedural, and world knowledge, FMSLTs can unlock predictive potential hidden in structured data ecosystems and enable autonomous agents to reason over complex workflows. \textbf{The choice is stark: continue pursuing correlation-based pattern matching, or build foundation models that understand the operational logic that gives data meaning~\cite{metaCWM2025,lehrach2025codeworldmodels}. We believe the future depends on choosing the latter.}

\typeout{FINAL_MAIN_BODY_PAGE \thepage}
\clearpage
{\small
\setlength{\bibsep}{0pt}
\bibliography{example_paper}
\bibliographystyle{icml2026}
}

\newpage
\appendix
\onecolumn

\definecolor{AppendixGray}{RGB}{55, 65, 81}
\definecolor{AppendixLightGray}{RGB}{249, 250, 251}
\definecolor{AppendixBorder}{RGB}{209, 213, 219}

\section*{Appendix}
\label{appendix:supply_chain}

\noindent\textit{Supplementary Materials for Supply-Chain Example}

\vspace{0.5em}
\noindent This appendix provides supplementary details for the supply-chain example introduced in Section~\ref{sec:intro} (with associated schematic illustration and multi-table schema in Fig.~\ref{fig:slt_ex} and Fig.~\ref{fig:manu_tab}), which were omitted from the main paper due to space constraints. Specifically, we present an acronym legend, descriptions of table functions, entity definitions, mockup configuration validation logic, and basic configuration rules.

\subsection*{A.1\quad Acronym Legend}

\begin{tcolorbox}[
  colback=AppendixLightGray,
  colframe=AppendixBorder,
  arc=2pt,
  boxrule=0.5pt,
  left=10pt,
  right=10pt,
  top=8pt,
  bottom=8pt
]
\begin{multicols}{2}
\small
\begin{itemize}[leftmargin=*, itemsep=3pt]
    \item\textbf{PRD}: Product (Core product definitions)
    \item\textbf{PROPT}: Product Option (Configurable product features)
    \item\textbf{PROPT\_VAL}: Product Option Value (Specific option choices)
    \item\textbf{CMP}: Component (Manufacturing components)
    \item\textbf{CFG\_CMP\_CN}: Configuration-Component Connection (Links options to components)
    \item\textbf{RMAT}: Raw Material (Base manufacturing materials)
    \item\textbf{BOM}: Bill of Materials (Component material requirements)
    \item\textbf{INV\_LOC}: Inventory Location (Storage facilities tracking)
    \item\textbf{INV\_TRX}: Inventory Transaction (Stock movement records)
    \item\textbf{SUPP}: Supplier (Vendor management)
    \item\textbf{PORD}: Purchase Order (Material procurement)
    \item\textbf{SHPMT}: Shipment (Delivery tracking)
    \item\textbf{CUST}: Customer (Client information)
    \item\textbf{ORD}: Order (Sales transactions)
    \item\textbf{ORD\_ITM}: Order Item (Per-product order details)
    \item\textbf{ORD\_CFG}: Order Configuration (Customer-chosen options)
\end{itemize}
\end{multicols}
\end{tcolorbox}

\subsection*{A.2\quad Table Functional Descriptions}

\begin{tcolorbox}[
  colback=AppendixLightGray,
  colframe=AppendixBorder,
  arc=2pt,
  boxrule=0.5pt,
  left=10pt,
  right=10pt,
  top=8pt,
  bottom=8pt
]
\begin{multicols}{2}
\small
\begin{itemize}[leftmargin=*, itemsep=3pt]
    \item \textbf{PRD}: Stores base product information and pricing
    \item \textbf{PROPT}: Defines configurable options for products (e.g., color, size)
    \item \textbf{PROPT\_VAL}: Contains specific option choices with price modifiers
    \item \textbf{CMP}: Tracks manufacturing components used in product assembly
    \item \textbf{CFG\_CMP\_CN}: Maps product configurations to required components
    \item \textbf{RMAT}: Manages raw materials inventory and suppliers
    \item \textbf{BOM}: Specifies raw material requirements for components
    \item \textbf{INV\_LOC}: Records physical storage locations and types
    \item \textbf{INV\_TRX}: Logs all inventory movements and adjustments
    \item \textbf{SUPP}: Maintains supplier contact and lead time information
    \item \textbf{PORD}: Tracks material procurement orders and status
    \item \textbf{SHPMT}: Monitors delivery status of purchased materials
    \item \textbf{CUST}: Stores customer personal information and contact details
    \item \textbf{ORD}: Manages order headers and overall status
    \item \textbf{ORD\_ITM}: Records line items with quantities and final pricing
    \item \textbf{ORD\_CFG}: Stores customer-selected configuration options per item
\end{itemize}
\end{multicols}
\end{tcolorbox}

\subsection*{A.3\quad Entity Definitions}

\newcommand{\entitytable}[3]{%
  \subsubsection*{\texttt{#1} --- #2}
  \vspace{0.3em}
  \noindent
  \begin{tabularx}{\textwidth}{>{\ttfamily\small}l>{\ttfamily\small}lX}
  \toprule
  \normalfont\bfseries Field & \normalfont\bfseries Type & \normalfont\bfseries Description \\
  \midrule
  #3
  \bottomrule
  \end{tabularx}
  \vspace{0.8em}
}

\entitytable{PRD}{Product}{
prd\_id & int & Unique product identifier (auto-increment) \\
prd\_name & varchar(255) & Product display name \\
prd\_desc & text & Detailed product description \\
base\_price & decimal(10,2) & Base price before configuration \\
created\_at & datetime & Timestamp of product creation \\
}

\entitytable{PROPT}{Product Option}{
propt\_id & int & Unique option identifier \\
prd\_id & int & Reference to PRD.prd\_id \\
opt\_name & varchar(100) & Option name (e.g., "Color") \\
display\_order & int & UI display sequence \\
}

\entitytable{PROPT\_VAL}{Product Option Value}{
prov\_id & int & Unique value identifier \\
propt\_id & int & Reference to PROPT.propt\_id \\
val\_name & varchar(100) & Value name (e.g., "Red") \\
price\_mod & decimal(10,2) & Price modifier for this value \\
}

\entitytable{BOM}{Bill of Materials}{
bom\_id & int & Unique BOM entry identifier \\
cmp\_id & int & Reference to CMP.cmp\_id \\
rmat\_id & int & Reference to RMAT.rmat\_id \\
qty\_req & decimal(10,2) & Required material quantity \\
}

\entitytable{INV\_TRX}{Inventory Transaction}{
trx\_id & int & Unique transaction ID \\
cmp\_id & int & Nullable component reference \\
rmat\_id & int & Nullable material reference \\
loc\_id & int & Reference to INV\_LOC.loc\_id \\
qty & decimal(10,2) & Transaction quantity \\
trx\_type & varchar(50) & Transaction type (PURCH/PROD/SALE/ADJ) \\
trx\_date & datetime & Transaction timestamp \\
notes & text & Additional transaction details \\
}

\entitytable{ORD\_CFG}{Order Configuration}{
ocfg\_id & int & Unique configuration ID \\
oitm\_id & int & Reference to ORD\_ITM.oitm\_id \\
prov\_id & int & Reference to PROPT\_VAL.prov\_id \\
}

\subsection*{A.4\quad Configuration Validation Logic}

\begin{tcolorbox}[
  colback=AppendixLightGray,
  colframe=AppendixBorder,
  arc=2pt,
  boxrule=0.5pt,
  title={\textbf{Algorithm: Computer Configuration Validator}},
  fonttitle=\bfseries,
  left=10pt,
  right=10pt,
  top=8pt,
  bottom=8pt
]
\begin{algorithmic}[1]
\STATE \textbf{Input:} User configuration $C$, Purpose $P$
\STATE \textbf{Output:} Valid configuration or error message

\STATE \textit{Apply purpose rules:}
\IF{$P == \text{``training''}$}
    \STATE $C.\text{GPU} \gets \text{PR.H200}$
\ELSIF{$P == \text{``inference''}$}
    \STATE $C.\text{GPU} \gets \text{PR.A10G}$
\ENDIF

\STATE \textit{Check physical compatibility:}
\IF{$C.\text{CPU.socket} \neq C.\text{Motherboard.socket}$}
    \STATE \textbf{return} Error: "CPU-Motherboard socket mismatch"
\ENDIF

\STATE \textit{Check power requirements:}
\STATE $total\_power \gets C.\text{CPU.TDP} + C.\text{GPU.TDP} + 100$ \COMMENT{100W buffer}
\IF{$total\_power > C.\text{PSU.wattage}$}
    \STATE \textbf{return} Error: "Insufficient PSU capacity"
\ENDIF

\STATE \textit{Check interface compatibility:}
\IF{$C.\text{GPU.interface} \neq C.\text{Motherboard.PCIe\_version}$}
    \STATE \textbf{return} Error: "GPU interface mismatch"
\ENDIF

\STATE \textit{Check RAM compatibility:}
\IF{$C.\text{RAM.type} \neq C.\text{Motherboard.RAM\_type}$}
    \STATE \textbf{return} Error: "RAM type not supported"
\ENDIF

\STATE \textbf{return} "Configuration valid"
\end{algorithmic}
\end{tcolorbox}

\subsection*{A.5\quad Purpose-Based Configuration Rules}

\begin{table}[h]
\centering
\begin{tabular}{lll}
\toprule
\textbf{Purpose} & \textbf{GPU Requirement} & \textbf{Minimum Specs} \\
\midrule
Training & PR.H200 & 32GB VRAM, PCIe 5.0 \\
Inference & PR.A10G & 24GB VRAM, PCIe 4.0 \\
Gaming & User choice & PCIe 4.0+ \\
Workstation & Quadro series & ECC memory support \\
\bottomrule
\end{tabular}
\caption*{\textbf{Table:} Purpose-based GPU configuration requirements}
\end{table}

\end{document}